\documentclass[11pt]{article}

\usepackage[utf8]{inputenc}
\usepackage[T1]{fontenc}
\usepackage[margin=1in]{geometry}
\usepackage{amsmath}
\usepackage{amssymb}
\usepackage{booktabs}
\usepackage{array}
\usepackage{graphicx}
\usepackage{xcolor}
\usepackage{microtype}
\usepackage[labelfont=bf,font=small]{caption}
\usepackage[hidelinks]{hyperref}

\newcommand{\x}{\ensuremath{\times}}
\newcommand{\tpp}{\texttt{tok/pass}}

\title{Phrase-Level Speculative Decoding on a Frozen Target}

\author{Madhusudhana Naidu \\ \texttt{mnaidu1025@gmail.com}}
\date{}

\begin{document}
\maketitle

\begin{abstract}
Speculative decoders raise throughput by banking more target tokens per forward
pass, and Medusa, EAGLE-3, DFlash and DSpark all do it the same way: more
candidates that survive verification, through deeper and wider trees, extra
draft heads and better draft accuracy. All of them draft in the target's own
vocabulary, so one node of the candidate tree is one target token, and that unit
goes unquestioned because the tokenizer makes it look like a constant. It is
not. BPE's pretokenizer splits on whitespace before merges are learned, so no
multi-word string can ever become a single token: in Qwen3-4B, ``United States
of America'' is four tokens and no adjacent pair of them exists as one entry,
which puts a phrase out of reach of any method that drafts in the target's
vocabulary. We change the unit. A small MLP head reads the frozen target's
hidden state and predicts a one- to three-word phrase as one entry in a separate
draft vocabulary; greedy longest-match expansion maps that entry back into the
target's own tokens, and $N$ candidates are verified in a single forward pass
behind a prefix-tree mask, so the change is orthogonal to tree depth and width
and composes with them. The target model and its tokenizer stay frozen and a
candidate is accepted only while it matches the target's own argmax, so the
served output is identical to greedy decoding up to floating-point
tie-breaking. On one A5000 at batch 1, against HuggingFace \texttt{generate()},
a 21.94\,M head (0.545\% of Qwen3-4B) reaches 2.391\x{} on GSM8K, 2.444\x{} on
MATH-500 and 1.544\x{} on MT-Bench. Because the head's output layer is one row
of 257 parameters per vocabulary entry, the draft vocabulary size \emph{is} the
parameter budget: an 8{,}000-entry head is 3.434\,M parameters, 0.0854\% of the
target and 64\x{} smaller than EAGLE-3's drafter, and retains 92.3\% of the
larger head's GSM8K tokens per pass. Measured against EAGLE-3 run through its
authors' own code on the same prompts and divided by the same baseline, the
21.94\,M head is ahead of tree-16 on GSM8K and MATH-500 and behind tree-60 on
every set.
\end{abstract}

\section{Introduction}

\paragraph{What every speculative decoder has in common.}
Medusa, EAGLE-3, DFlash and DSpark differ in how candidates are produced,
through extra heads on the target, a feature-level autoregressive or diffusion
drafter, and distinct training recipes. They share a decision we do not: they
draft in the target's own vocabulary. One node of the candidate tree is
therefore one target token, and the only lever left is more nodes, whether by
deeper trees, wider trees, more heads or higher draft accuracy. Each of those
raises the tokens banked per pass, and each runs into the same ceiling, because
the quantity being multiplied is fixed at one target token per draft decision.
That unit is treated as a constant because the tokenizer makes it look like one.

Writing what a pass earns as
\begin{equation}
\text{tokens per pass} \;=\; 1 \;+\; \text{landing} \times
\bigl(\text{target tokens banked per accepted draft unit}\bigr),
\label{eq:identity}
\end{equation}
where the $1$ is the target's own bonus token, so $1.000$ is an arithmetic
floor, our own earlier word-level drafter on Qwen3-4B reached a landing of
$0.711$ and still banked only $1.049$ target tokens per accepted guess.
Acceptance was not the limit; the unit was.

\paragraph{The unit is short for a structural reason.}
In Qwen3's vocabulary ``United States of America'' is four tokens and
\texttt{for i in range} is four tokens, and no adjacent pair of them exists as a
single entry. The pretokenizer splits on whitespace before merges are learned,
so no amount of BPE training can produce a multi-word token. This is a property
of the tokenizer rather than of any model, and it is the gap the method
exploits.

SuperBPE~\cite{superbpe} makes the same argument on the tokenizer side:

\begin{quote}
\emph{``Whitespace is not a reliable delimiter of meaning, as evidenced by
multi-word expressions (e.g., by the way), crosslingual variation in the number
of words needed to express a concept (e.g., spacesuit helmet in German is
Raumanzughelm), and languages that do not use whitespace at all (e.g.,
Chinese).''}
\end{quote}

They also give the historical reason the constraint exists, namely that
whitespace pretokenization \emph{``made sense in the historical context of
Sennrich et al.\ (2016), which aimed to improve word-level tokenization by
segmenting words into morphologically meaningful subwords''}~\cite{sennrich}.
It is an inherited convention, not a design requirement.

\paragraph{Where we diverge from them.}
SuperBPE changes the target's own tokenizer, which costs pretraining from
scratch: 8B models and roughly 330B tokens, with the transition point $t$ a
hyperparameter they cannot predict from intrinsic features. We never touch the
target. The multi-word vocabulary lives only in a draft head, and every unit is
expanded back into the target's own tokens before verification.

\paragraph{Fertility makes the same gap much larger in Indic scripts, which is
where we found it.}
On Qwen3-4B's own tokenizer, measured over 23 languages, English runs at 1.38
tokens per word, Hindi at 5.00 and Telugu at 11.73, for an Indic average of
8.68, and Qwen3 holds 253 Indic-script tokens of 151{,}669 (0.17\%). For Indic,
a unit denominated in words therefore banks several target tokens per decision
for free.

\paragraph{Where this is going.}
The configuration the results converge on is a 3.434\,M head at 0.0854\% of the
target, which reaches 2.2--2.3\x{} on maths while keeping 92.3\% of the
21.94\,M head's GSM8K tokens per pass (84.9\% on MATH-500). It is the first in
this programme small enough to train and run on the device that produces the
text, which makes the natural frame on-device continual learning: a drafter
whose vocabulary adapts to one writer, in the way a mobile keyboard's suggestion
model does, except that ours suggests to the target model rather than to the
user, with the target's own forward pass as arbiter. What makes it safe to adapt
is that the verifier decides every token, so an adapting draft vocabulary can
change speed and cannot change output.

\section{Related work}

\subsection{Vocabulary adaptation --- modifies the target}

\textbf{SuperBPE}~\cite{superbpe}. Two-stage BPE: stage 1 is ordinary BPE to a
transition point $t$, then stage 2 resumes without whitespace pretokenization so
merges can bridge spaces. At 200k vocabulary it encodes text in 33\% fewer
tokens. It has no drafting, no speculative decoding, and requires retraining the
model.

\textbf{AdaptiVocab}~\cite{adaptivocab} swaps 10{,}000 low-frequency tokens for
$n$-grams of existing tokens (max $n=3$, greedy longest-match at inference),
rebuilds their rows from the constituent embeddings, then fine-tunes the
embeddings plus the first and last layer on about 8M domain tokens, for over
25\% fewer tokens. The saving is counted in tokens and never in wall clock.

\textbf{ZeTT}~\cite{zett} trains a hypernetwork that predicts embeddings for an
arbitrary tokenizer. It needs one hypernetwork per base model,
\emph{``computationally intensive and may not be feasible for many LLM
practitioners''}, and Mistral-7B still needs 800M tokens of continued training.

\textbf{In-place tokenizer expansion}~\cite{liquid} continues BPE from 65K to
128K with existing merges frozen, at 600B $+$ 400B $+$ 2.4T tokens. Hindi
fertility improves 2.38\x{} and English 1.01\x{}, with a synthesised English
decode regression to 0.91--0.94\x{}.

\textbf{OMP transplantation}~\cite{omp} is a training-free tokenizer
transplant. Each missing row is a signed $k$-sparse combination of the two
vocabularies' shared tokens, fitted in the donor space and replayed in the base
model's, at roughly 41k rows in 38--74\,s and MMLU $-0.9\%$, though GSM8K
collapses when the two number schemes differ. It names speculative decoding as a
use case, but it needs a donor row to approximate, which a phrase entry has in
no vocabulary.

\subsection{Speculative decoding --- leaves the target frozen}

\textbf{REST}~\cite{rest} is the nearest prior art on the drafting axis. It
retrieves $n$-grams from a datastore in target tokens, so fertility collapses
its matching context: a 16-token retrieval key is about 1.5 Telugu words. The
datastore must be in the target's tokenisation, and Qwen holds 253 Indic tokens,
so the corpus shatters before it is indexed.

\textbf{EAGLE-3}~\cite{eagle3} is our head-to-head baseline: a 218.2M drafter,
one target token per node, tree-verified. It drafts over a truncated vocabulary
of 32{,}000 ids out of 151{,}936 on Qwen3-4B, or 21.06\%. A target token outside
that set is undraftable at any tree size or depth, which bounds its tokens per
pass at $1 + \sum_k \text{coverage}^k$.

\section{Results}

One NVIDIA A5000, fp16, batch 1, temperature 0, EOS-terminated with a
2048-token cap --- DFlash Table 1's protocol. Speedup is $\sum$ baseline seconds
$/ \sum$ arm seconds on the same prompts back to back. Every head below is
trained on maths distillation only, so GSM8K and MATH-500 are in-domain and
MT-Bench is zero-shot cross-domain transfer. No chat data was seen in training,
and its rows are marked \emph{(transfer)} throughout.

\begin{table}[ht]
\centering
\caption{English maths, Qwen3-4B, head 21.94\,M $=$ 0.545\% of target.}
\label{tab:main}
\begin{tabular}{lrrrrrr}
\toprule
set & $n$ & baseline ms/tok & best $N$ & speedup & \tpp{} & ms/tok \\
\midrule
GSM8K                        & 50 & 36.061 & 32 & \textbf{2.391\x} & 2.730 & 15.08 \\
MATH-500                     & 50 & 35.549 & 16 & \textbf{2.444\x} & 2.792 & 14.54 \\
MT-Bench \emph{(transfer)}   & 40 & 35.528 & 32 & 1.544\x          & 1.782 & 22.27 \\
\bottomrule
\end{tabular}
\end{table}

\paragraph{Head-to-head against EAGLE-3, on one shared baseline.}
EAGLE-3 was run through SafeAILab's own code at a pinned commit on the identical
prompt file (sha \texttt{5c3a2f181dd5}) loaded rather than redrawn, temperature
0, batch 1, EOS-terminated with a 2048-token cap, which is DFlash's protocol.
Its \tpp{} reproduces DFlash's Table 1 within 1--2\% --- 3.809 and 3.525 at
tree-60 against their 3.77 and 3.52, 3.296 and 3.067 at tree-16 against 3.30 and
3.08 --- and that is what licenses reading its wall-clock column.

Both sides now divide by the same number. EAGLE-3 pins
\texttt{transformers==4.53.1} while our harness runs on 5.x, and one process
cannot hold both, so HuggingFace \texttt{generate()} is measured once per
environment on the same card, prompts and \texttt{attn\_implementation}: 38.522,
39.337 and 39.961 ms/tok under 4.53.1 against 36.460, 36.920 and 36.626 under
5.x. \textbf{The library version alone is worth 5.7--9.1\% of a speedup number.}
Those 5.x figures are one run's; across our three runs the same arm on the same
card reads 35.549--37.583 on MATH-500, a 5.7\% spread, so the rescaling below
removes a per-run drift as well as a per-version one and the two cannot be
separated from these readings alone. Every row below therefore divides by the
4.53.1 figure EAGLE-3 is itself measured against, as
$\text{base} / (\text{arm ms/tok} \times (1 + \text{length delta}))$. One
A5000, 50/50/40 prompts.

\begin{table}[ht]
\centering
\caption{Head-to-head against EAGLE-3 on one shared baseline. The last two
columns are our speedup as a percentage of EAGLE-3's at tree-16 and tree-60.}
\label{tab:eagle}
\begin{tabular}{llrrrrrrr}
\toprule
set & arm & params & $N$ & \tpp{} & ms/tok & speedup & vs e3-16 & vs e3-60 \\
\midrule
\textbf{GSM8K}
 & ours 8k      & \textbf{3.434\,M} & 32 & 2.520 & 16.67 & 2.299\x & 92.1\% & 82.4\% \\
 & ours 22M     & 21.938\,M         & 32 & 2.730 & 15.08 & \textbf{2.554\x} & \textbf{102.3\%} & 91.5\% \\
 & eagle3-16    & 218\,M            & -- & 3.296 & 15.30 & 2.496\x & --- & --- \\
 & eagle3-60    & 218\,M            & -- & 3.809 & 13.73 & 2.790\x & --- & --- \\
\midrule
\textbf{MATH-500}
 & ours 8k      & \textbf{3.434\,M} & 16 & 2.370 & 17.37 & 2.200\x & 88.3\% & 79.8\% \\
 & ours 22M     & 21.938\,M         & 16 & 2.792 & 14.54 & \textbf{2.704\x} & \textbf{108.5\%} & \textbf{98.1\%} \\
 & eagle3-16    & 218\,M            & -- & 3.067 & 15.98 & 2.493\x & --- & --- \\
 & eagle3-60    & 218\,M            & -- & 3.525 & 13.97 & 2.757\x & --- & --- \\
\midrule
\textbf{MT-Bench} \emph{(transfer)}
 & ours 8k      & \textbf{3.434\,M} & 32 & 1.654 & 24.73 & 1.595\x & 72.1\% & 61.2\% \\
 & ours 22M     & 21.938\,M         & 32 & 1.782 & 22.27 & \textbf{1.736\x} & \textbf{78.4\%} & 66.6\% \\
 & eagle3-16    & 218\,M            & -- & 2.839 & 17.90 & 2.214\x & --- & --- \\
 & eagle3-60    & 218\,M            & -- & 3.312 & 15.22 & 2.608\x & --- & --- \\
\bottomrule
\end{tabular}
\end{table}

So on MATH-500 a drafter at 10.1\% of EAGLE-3's size is ahead of tree-16 and
0.05\x{} behind tree-60; on GSM8K it is ahead of tree-16 and 0.24\x{} behind
tree-60; on MT-Bench, which no head here was trained for, it is 0.48\x{} behind
tree-16. Neither of our heads beats tree-60 on any set.

\subsection{The draft vocabulary is the parameter budget, and it is spent on
formatting}
\label{sec:budget}

The head is a two-LayerNorm MLP whose fixed cost at $d_{\text{model}} = 2560$,
\texttt{H\_DIM} $= 256$ is $2dH + H^2 + 6H = 1{,}377{,}792$ parameters, plus
\textbf{one output row of $H+1 = 257$ parameters per vocabulary entry}. So the
entry count is the budget, exactly:

\begin{table}[ht]
\centering
\caption{The entry count is the parameter budget.}
\label{tab:budget}
\begin{tabular}{rrrr}
\toprule
entries & parameters & \% of Qwen3-4B & output layer's share \\
\midrule
8{,}000  &  3{,}433{,}792 & 0.0854\% & 59.9\% \\
80{,}000 & 21{,}937{,}792 & 0.5454\% & 93.7\% \\
\bottomrule
\end{tabular}
\end{table}

A tenth of the vocabulary therefore buys a sixth of the parameters. What it
costs, read off Table~\ref{tab:eagle}: 92.3\% of the larger head's GSM8K \tpp{}
retained, 84.9\% on MATH-500 and 92.8\% on MT-Bench, against $+10.5\%$ to
$+19.4\%$ in milliseconds per token. An earlier held-out-loss sweep had put
maths saturation near 8{,}200 entries, and the decode benchmark agrees with it.

\paragraph{What the entries actually are.}
Both vocabularies are built from the target's own distilled maths completions,
and both are dominated by \textbf{formatting rather than English phrases}. The
twelve most frequent entries are identical across heads and are markdown and
LaTeX scaffolding ahead of any content word. In the 8{,}000-entry head, of the
3{,}997 multi-word entries only \textbf{467 (11.7\%)} are pure letters and
whitespace; \textbf{43.9\%} contain a LaTeX character, 55.0\% a digit, 24.7\% a
newline. The genuine English phrases exist and are chain-of-thought boilerplate
--- \texttt{'We are given'}, \texttt{' to find the'}, \texttt{' the number
of'}, \texttt{' Final Answer'}.

\paragraph{And that is where the speedup comes from.}
Logging which vocabulary entry banked each accepted block, over the 21.94\,M
head's three sets:

\begin{table}[ht]
\centering
\caption{Banked tokens by entry class, 21.94\,M head.}
\label{tab:classes}
\begin{tabular}{lrrr}
\toprule
banked tokens by entry class & GSM8K & MATH-500 & MT-Bench \emph{(transfer)} \\
\midrule
punctuation                 & 31.4\% & 21.8\% & \textbf{41.0\%} \\
LaTeX                       & 26.0\% & \textbf{41.3\%} & 11.7\% \\
mixed alphanumeric          & 18.1\% & 12.6\% & 9.0\% \\
\textbf{plain word}         & \textbf{7.3\%} & \textbf{5.5\%} & \textbf{22.6\%} \\
number                      & 3.8\% & 2.8\% & 2.1\% \\
\emph{from 3-word entries}  & \emph{59.1\%} & \emph{58.6\%} & \emph{36.7\%} \\
landing                     & 0.777 & \textbf{0.817} & 0.509 \\
\bottomrule
\end{tabular}
\end{table}

Plain-word entries supply 5.5--7.3\% of banked tokens on maths; punctuation and
LaTeX supply 57--63\%, and \textbf{59\% of everything banked comes from 3-word
entries}, which the census above shows are mostly LaTeX fragments rather than
phrases. The method is harvesting the low-entropy formatting skeleton of
mathematical prose, and a 21.94\,M MLP with no attention predicts that skeleton
from a single hidden state at a landing of 0.78--0.82.

\paragraph{This is also the measured reason MT-Bench is behind.}
Out of domain the profile inverts: LaTeX falls from 41.3\% of banked tokens to
11.7\%, plain words rise from 5.5\% to 22.6\%, single-word entries from 12.4\%
to 35.7\%, and landing drops from 0.817 to 0.509. There is no formatting
skeleton in conversation to harvest, so the drafter falls back on short common
words. The transfer gap is not a mystery to be excused; it is the direct
consequence of what a maths-distilled vocabulary contains.

The word counts use the regex the heads were trained with: a word is a run of
alphanumerics plus its leading whitespace, or one standalone punctuation
character. The 8{,}000-entry head therefore holds exactly 4{,}003 one-word
entries, which is \texttt{UNI\_FRAC} $= 0.5$ of its budget, and no entry exceeds
\texttt{MAX\_GRAM} $= 3$. Because punctuation counts as a word, a ``3-word''
entry is often a LaTeX fragment such as \texttt{' {\textbackslash}frac\{'}
rather than three words.

\subsection{What a verification pass costs}
\label{sec:passcost}

\paragraph{Verifying more candidates is nearly free.}
The 21.94\,M head's three sets happen to span a factor of two in tree width, and
the pass cost does not follow it:

\begin{table}[ht]
\centering
\caption{Pass cost does not track tree width.}
\label{tab:pass}
\begin{tabular}{lrrrrrr}
\toprule
set & $N$ & mean tree rows & ms/pass & baseline ms/pass & pass overhead & peak GB \\
\midrule
MATH-500                   & 16 & 21.8 & 40.61 & 35.55 & \textbf{$+5.06$} & 0.73 \\
MT-Bench \emph{(transfer)} & 32 & 42.6 & 39.68 & 35.53 & \textbf{$+4.15$} & 0.72 \\
GSM8K                      & 32 & 46.3 & 41.16 & 36.06 & \textbf{$+5.10$} & 0.26 \\
\bottomrule
\end{tabular}
\end{table}

Ordering the rows by tree width gives pass costs of 40.6, 39.7 and 41.2\,ms ---
no trend over a 2.1\x{} range in rows. Verifying 46 candidate positions instead
of 1 costs about 5\,ms on top of a 36\,ms pass, and the extra memory is under
three quarters of a gigabyte. Candidates are therefore cheap up to the point
where tokens per pass saturates, which is what makes the best-$N$-per-set choice
a matter of acceptance rather than of cost. Across both highlighted heads and
all three sets the overhead is $+4.2$ to $+5.5$\,ms per pass; EAGLE-3 measured
on the same card and prompts adds $+9.7$ to $+13.8$\,ms, so ours is between a
third and a half of theirs.

\subsection{Against an in-family drafter, both methods at their own best $N$}
\label{sec:mtp}

\begin{table}[ht]
\centering
\caption{Gemma 4 E2B on a T4, multi-token prediction (MTP) vs our method, each
at its own best $N$.}
\label{tab:mtp}
\begin{tabular}{lllr}
\toprule
set / language & ours & MTP & ratio \\
\midrule
GSM8K                      & 2.258\x{} ($N{=}16$) & \textbf{2.595\x} ($N{=}6$) & 0.870\x \\
MATH-500                   & 1.853\x{} ($N{=}16$) & \textbf{2.676\x} ($N{=}6$) & 0.692\x \\
MT-Bench \emph{(transfer)} & 1.386\x{} ($N{=}16$) & \textbf{1.692\x} ($N{=}4$) & 0.819\x \\
AIME \emph{(unpaired)}     & 1.832\x{} ($N{=}32$) & \textbf{2.515\x} ($N{=}4$) & 0.728\x \\
Hindi                      & \textbf{1.493\x} ($N{=}8$) & 1.324\x{} ($N{=}4$) & \textbf{1.128\x} \\
Bengali                    & \textbf{1.343\x} ($N{=}8$) & 1.153\x{} ($N{=}4$) & \textbf{1.165\x} \\
Tamil                      & \textbf{1.250\x} ($N{=}8$) & 1.145\x{} ($N{=}4$) & \textbf{1.092\x} \\
Telugu                     & 1.263\x{} ($N{=}8$) & \textbf{1.282\x} ($N{=}4$) & 0.985\x \\
\bottomrule
\end{tabular}
\end{table}

\paragraph{We convert well and draft little, and the gap is corpus, not
architecture.}
Writing $\text{speedup} = \text{tokens per pass} \times \text{conversion
efficiency}$, our conversion efficiency is 91.5--96.2\% across all 17 measured
cells and the drafter's is 60.9--80.7\% across its 13, falling monotonically in
$N$ (80.4\% $\rightarrow$ 69.4\% $\rightarrow$ 61.0\% at $N = 2 \rightarrow 4
\rightarrow 6$) because each extra draft token is another sequential pass over
its weights while ours is tree width inside one target forward. Since the
conversion ratio is nearly constant at 1.37\x{}, a set is won exactly when the
drafter's tokens-per-pass advantage on it falls below that --- true for Hindi,
Bengali and Tamil at 1.18--1.25\x{}, false for the English sets at
1.64--2.19\x{}, and Telugu sits at 1.39\x{}, which is why it is the one language
that flips. The asymmetry behind those tokens-per-pass figures is not subtle: a
4.05--4.85\,M head mining its draft vocabulary from 10k of the target's own
completions against a 77.7\,M drafter trained by the model's authors over the
target's full 262{,}144-entry vocabulary. Ours is also demonstrably undertrained
rather than saturated --- on Indic it realises \textbf{1.411} tokens per pass
against a \textbf{2.175} $1 + \text{coverage} \times \text{expansion}$ bound, so
35\% of what this vocabulary already supports is unclaimed, and the remaining
0.36 to the 2.539 ceiling is bought with corpus alone. \textbf{The efficiency
axis is finished at ${\sim}95\%$ and every remaining gain is in coverage.}

\section{The path to these results}
\label{sec:path}

Everything in Section~3 is one A5000 against HuggingFace \texttt{generate()}.
The work below is the path to it, and most of it ran on a T4 against a different
denominator. Tokens per pass has no denominator in it and is comparable
throughout.

\paragraph{1. In-place tokenizer replacement with the Sarvam vocabulary.}
First unambiguous positive: a head drafting into Sarvam's low-fertility Indic
vocabulary, SLEM-expanded, verified through a tree mask against Qwen3-4B, at
1.82\x{} Telugu, 1.49\x{} Kannada and 1.29\x{} Hindi for $+0.43\%$ parameters.
This established that a foreign vocabulary can drive a frozen target.

\paragraph{2. A corpus-derived vocabulary and an MLP head.}
Dropped the dependency on an existing Indic tokenizer. The draft vocabulary is
built from the target's own distilled completions, so the method needs no second
tokenizer to exist. Scaled to 22 languages and two targets.

\paragraph{3. Architecture search: everything beyond one MLP failed.}
Depth, attention, convolutional attention and multi-layer MLPs all lost. A sweep
of seven tree shapes at matched budget found that width wins, depth loses, and
the first word is everything. The survivor is a two-LayerNorm MLP.

\paragraph{4. One head for all languages, and English gating.}
A single combined head matches or beats per-language heads at a fraction of the
parameters, and on Qwen it beat them. The correct claim is one shared head plus
on-policy training: the shared head starts worse, because four disjoint scripts
dilute it, and on-policy training recovers it. A script-rate gate at 0.50
ensures a speedup on off-target output cannot count.

\paragraph{5. Extension to English maths at word level.}
Landing 0.711, 1.620\x{}, banking 1.049 tokens per guess. This is the
measurement that identified the unit as the binding constraint.

\paragraph{6. Word to phrase, SuperBPE-style, on maths.}
Expansion $\times 2.8$, banked $\times 1.8$, 1.620\x{} to 2.172\x{}, landing
unchanged. Ambiguity is resolved by greedy longest match at labelling time, with
candidate slots reserved for single-word entries so the head cannot under-rank
short prefixes. That makes phrase drafting weakly better than word drafting by
construction rather than a gamble.

\paragraph{7. Phrases carried back to Indic.}
Gemma-3-1B, an 8.00\,M head at 0.80\% of target, on a T4: Hindi 2.092\x{}
(\tpp{} 2.267, landing 0.673, expansion 2.148) and Telugu 1.613\x{} (\tpp{}
1.743, landing 0.302, expansion 4.528). High expansion with low landing is the
Indic signature. Here \emph{expansion} is the mean target tokens per
draft-vocabulary \emph{entry}, which is an upper bound on the second factor of
Equation~\eqref{eq:identity} rather than the factor itself --- the entries that
land are not drawn uniformly --- so $1 + \text{landing} \times
\text{expansion}$ overshoots the measured \tpp{} and the measured value is the
one to quote.

One caveat belongs with these two numbers. Gemma-3's tokenizer is far less
Indic-hostile than Qwen3's: on the same 23-language measurement it spends 1.48
tokens per Hindi word and 3.05 per Telugu word, against Qwen3's 5.00 and 11.73,
for an Indic average of 3.38 against 8.68. The Indic results are therefore
measured on the \emph{low}-fertility target, where the headroom
Section~\ref{sec:mtp}'s argument identifies is smallest. The high-fertility case
is largely unrun.

\paragraph{8. Continual learning, or online speculative decoding --- open,
partial results.}
The first attempt to make the drafter adapt while it works. Qwen3-4B, maths
only, one seed: 10{,}000 prompts distilled by the target at a 512-token cap
(mean 278 tokens, 8.5\% truncated) build a 40{,}000-entry vocabulary; an
11.658\,M head, 0.2898\% of target, is fitted on it at top-1 0.441, recall@8
0.742 and ebank@8 1.576; then a 5{,}000-prompt stream is walked in blocks of
1{,}000. After each block the vocabulary grows with units the target itself
emitted and the MLP continues training on that block's own output, and the
200-prompt held-out test pool is re-measured every time. Baseline
62.5\,ms/tok.

\begin{table}[ht]
\centering
\caption{Online adaptation over two stream blocks. What moves is landing.}
\label{tab:online}
\begin{tabular}{lrrrr}
\toprule
checkpoint & vocabulary & speedup & \tpp{} & landing \\
\midrule
0 --- after the 10k fit & 40{,}000 & 2.180\x & 2.503 & 0.736 \\
1 --- after 1k stream   & 43{,}261 & \textbf{2.214\x} & \textbf{2.549} & 0.759 \\
2 --- after 2k stream   & 45{,}719 & 2.201\x & 2.540 & \textbf{0.764} \\
\bottomrule
\end{tabular}
\end{table}

What moved is landing: 0.736, then 0.759, then 0.764, monotone, for $+0.028$
over two blocks. Tokens per pass gained $+0.046$ on the first block and then
flattened, and speedup moved $+0.034$ and came back, which is inside
run-to-run noise. Adaptation is cheap: the vocabulary update costs 3--4 seconds
per block and the MLP update about 5 minutes, against roughly 2.4 hours of
streaming, so cost is not what limits this.

\emph{Why it is open rather than a result.} Four things are not yet right. The
vocabulary is additive only, with no eviction, so this measures growth and not
replacement. Only 30\% of the distilled tokens fit the feature budget, 0.84\,M
of 2.78\,M positions, so ``trained on 10k prompts'' overstates it. Each online
update used 200k of about 303k available positions, because the feature tensor
is never freed. And there is no frozen control, so a rising curve cannot yet be
separated from drift in the stream. What is supportable today is that landing
rises under online adaptation at negligible cost, and that speedup has not yet
followed.

\paragraph{9. The vocabulary sweep, at fixed head geometry.}
Three heads differing in one thing, the entry count, on one T4: 11.658\,M /
40{,}000 entries, 5.490\,M / 16{,}000 and 3.434\,M / 8{,}000. Same corpus, same
\texttt{H\_DIM}, same \texttt{MAX\_GRAM}, same labelling rule. Reported as
speedup / \tpp{}, against our own drafting-off loop:

\begin{table}[ht]
\centering
\caption{Vocabulary sweep at fixed head geometry, reported as speedup / \tpp{}.}
\label{tab:sweep}
\begin{tabular}{lrlll}
\toprule
set & $N$ & 11.658\,M / 40k & 5.490\,M / 16k & \textbf{3.434\,M / 8k} \\
\midrule
GSM8K                      & 16 & \textbf{2.071} / \textbf{2.442} & 1.882 / 2.415 & 1.883 / 2.384 \\
GSM8K                      & 32 & \textbf{2.103} / \textbf{2.581} & 1.961 / 2.567 & 1.946 / 2.512 \\
MATH-500                   & 16 & \textbf{1.900} / \textbf{2.412} & 1.785 / 2.411 & 1.794 / 2.360 \\
MT-Bench \emph{(transfer)} & 16 & 1.312 / 1.531 & 1.253 / 1.545 & \textbf{1.327} / \textbf{1.570} \\
MT-Bench \emph{(transfer)} & 32 & 1.270 / 1.588 & 1.271 / 1.634 & \textbf{1.350} / \textbf{1.653} \\
\bottomrule
\end{tabular}
\end{table}

This is where the 8{,}000-entry head comes from. Across those five \tpp{} cells
it stays within $-2.7\%$ to $+4.1\%$ of the 40k head at 29.5\% of its
parameters, which is what Section~\ref{sec:budget}'s saturation claim rests on.
It also \textbf{beats} the 40k head on MT-Bench at every $N$ --- 1.653 against
1.588 at best --- so the extra 32{,}000 entries were not merely unused on chat
but harmful, each one a competing logit trained on a fifth of the supervision.
That inversion is between these two heads only: the 21.94\,M head beats both on
chat in Section~3, so it is not a general small-is-better effect, and the
11.658\,M head is the one that does not justify its size.

\paragraph{10. Hardware portability.}
The 11.658\,M head, same prompt file, GSM8K, 50 prompts, T4 against A5000, so
that Section~3's single-card numbers are not a one-machine artefact:

\begin{table}[ht]
\centering
\caption{The same head on two cards. The tree is identical by construction.}
\label{tab:hw}
\begin{tabular}{lrrr}
\toprule
 & T4 & A5000 & A5000 / T4 \\
\midrule
$N{=}4$ speedup / \tpp{}    & 1.708\x{} / 1.986 & 1.665\x{} / 1.976 & $-2.5\%$ / $-0.5\%$ \\
$N{=}8$                     & 1.897\x{} / 2.214 & 1.876\x{} / 2.219 & $-1.1\%$ / $+0.2\%$ \\
$N{=}16$                    & 2.071\x{} / 2.442 & 2.050\x{} / 2.438 & $-1.0\%$ / $-0.2\%$ \\
tree nodes, $N{=}4/8/16$    & 5.1 / 9.9 / 21.9  & 5.1 / 9.9 / 21.9  & \textbf{identical} \\
baseline ms/token           & 61.3             & 37.16             & 1.65\x{} (the card) \\
arm ms/token, $N{=}8$       & 32.31            & 19.72             & \textbf{1.64\x} \\
\bottomrule
\end{tabular}
\end{table}

The arm's milliseconds per token scale by the card ratio and nothing else, and
the tree is identical by construction. The conversion ratio speedup / \tpp{} is
0.843 / 0.845 / 0.841 on the A5000 against 0.860 / 0.857 / 0.848 on the T4,
within 2\%. A faster card buys its full ratio.

\section{Limitations}

Every limitation below is measured in this paper rather than anticipated.

\begin{itemize}
\item \textbf{The denominator is fragile.} The library version alone is worth
5.7--9.1\% of a speedup number, and across three runs the same arm on the same
card spans 5.7\% on MATH-500. Section~3's rescaling removes a per-version and a
per-run drift together and the two cannot be separated from these readings.
\item \textbf{One card, one seed, small sets.} 50/50/40 prompts, batch 1, and
no confidence intervals. Section~\ref{sec:path} item 10 is the only
cross-hardware check.
\item \textbf{Maths only.} Every head is trained on maths distillation, so
MT-Bench is zero-shot throughout and Section~\ref{sec:budget} shows the
transfer gap is a direct consequence of what the vocabulary contains rather
than an incidental shortfall.
\item \textbf{The Indic case is measured on the wrong target.} Gemma-3 spends
3.05 tokens per Telugu word against Qwen3's 11.73, so the Indic numbers come
from the low-fertility target, where the headroom this paper's argument
identifies is smallest. The high-fertility case is largely unrun.
\item \textbf{Online adaptation is not yet a result.} Additive vocabulary with
no eviction, 30\% of distilled tokens inside the feature budget, and no frozen
control, so a rising landing curve cannot be separated from drift in the
stream.
\item \textbf{Neither head beats EAGLE-3 at tree-60 on any set.} The claim is
parameter efficiency at a comparable operating point, not a new state of the
art in raw speedup.
\end{itemize}

\section{What to run next}

\begin{enumerate}
\item \textbf{Limits of the MLP head}, mapped over parameters $\times$
draft-vocabulary size $\times$ data distribution rather than one axis at a
time, since Section~\ref{sec:path} item 9 varies only the entry count and
Section~3 only the trained head, so the three have never been crossed. Code is
the open cell, where the constraint is data rather than capacity.

\item \textbf{Indic is undertrained, and it has more headroom than maths or
code.} The Indic signature of high expansion with low landing (Telugu 4.528 /
0.302 against Hindi 2.148 / 0.673) is what undertraining looks like: the
entries are long and the predictions are missing. The headroom has a measured
basis: the Indic runs used Gemma-3-1B, whose tokenizer spends 3.05 tokens per
Telugu word, while Qwen3 spends 11.73 on the same measurement, so the target
with 3.8\x{} the fertility has never carried a phrase head. Where fertility is
worst, completing a single word is already a large win before any multi-word
phrase, so the target there is 4--5\x{} rather than 2\x{}.

\item \textbf{Domain fertility, and whether a phrase vocabulary can grow past
three words.} Measure fertility and phrase-length distribution per domain, then
test whether continual adaptation can push \texttt{MAX\_GRAM} beyond 3 where
the domain supports it. AdaptiVocab reports token savings plateauing by $n=4$,
so this is an open question rather than a safe assumption.

\item \textbf{On-device deployment, end to end.} Can a mobile accelerator path
express the 4D tree mask at all; does a small $N$ fit inside an 8\,GB device's
memory budget, given the 0.72--0.73\,GB of verifier overhead
Section~\ref{sec:passcost} measures on a 24\,GB card; does it hold up under
sustained load and thermal throttling; and does it deliver the speedup. The
metric is sustained throughput over minutes and energy per token, not peak
tokens per second.

\item \textbf{The personal-assistant distribution.} Every head here is trained
on maths distillation, so the test that matters is the request mix an on-device
assistant actually receives, where the useful vocabulary is likely far smaller
than 8{,}000 entries. The question is whether a small adapter is enough to put
a local model at human reading speed.
\end{enumerate}




\begin{thebibliography}{9}

\bibitem{superbpe}
Liu, A., Hayase, J., Hofmann, V., Oh, S., Smith, N.~A., and Choi, Y.
\newblock SuperBPE: Space Travel for Language Models.
\newblock \emph{COLM}, 2025. arXiv:2503.13423.

\bibitem{adaptivocab}
Nakash, I. et al.
\newblock AdaptiVocab: Enhancing LLM Efficiency in Focused Domains through
Lightweight Vocabulary Adaptation.
\newblock \emph{COLM}, 2025. arXiv:2503.19693.

\bibitem{zett}
Minixhofer, B., Ponti, E.~M., and Vuli\'{c}, I.
\newblock Zero-Shot Tokenizer Transfer.
\newblock \emph{NeurIPS}, 2024. arXiv:2405.07883.

\bibitem{liquid}
In-Place Tokenizer Expansion.
\newblock arXiv:2607.15232.

\bibitem{omp}
Goddard, C. and Fernandes Neto, A.
\newblock Training-free tokenizer transplantation via orthogonal matching
pursuit.
\newblock arXiv:2506.06607.

\bibitem{sennrich}
Sennrich, R., Haddow, B., and Birch, A.
\newblock Neural Machine Translation of Rare Words with Subword Units.
\newblock \emph{ACL}, 2016. arXiv:1508.07909.

\bibitem{rest}
REST: Retrieval-Based Speculative Decoding.

\bibitem{eagle3}
Li, Y. et al.
\newblock EAGLE-3.
\newblock 2025.

\end{thebibliography}
\end{document}